\documentclass[letterpaper, 10 pt, conference]{ieeeconf}  

\IEEEoverridecommandlockouts                              

\newcommand{\etal}{\textit{et al.}}
\usepackage{times}
\usepackage{epsfig}
\usepackage{graphicx}
\usepackage{amsmath}
\usepackage{amssymb}
\usepackage{hyperref}

\title{\LARGE \bf
Learning to Remove Multipath Distortions in Time-of-Flight Range Images for a Robotic Arm Setup}

\author{Kilho Son$^{1}$, \quad Ming-Yu Liu$^{2}$, \quad and Yuichi Taguchi$^{2}$
\thanks{*This work was fully supported by Mitsubishi Electric Research Labs (MERL), Cambridge, MA 02139, USA}
\thanks{$^{1}$Kilho Son is with Brown University,  Providence, RI 02912, USA
        {\tt\small kilho\_son@brown.edu }}%
\thanks{$^{2}$Ming-Yu Liu and Yuichi Taguchi are with Mitsubishi Electric Research Labs (MERL), Cambridge, MA 02139, USA
        {\tt\small \{mliu,taguchi\}@merl.com}}%
}

\begin{document}
\maketitle
\thispagestyle{empty}
\pagestyle{empty}

\begin{abstract}
Range images captured by Time-of-Flight (ToF) cameras are corrupted with multipath distortions due to interaction between modulated light signals and scenes. The interaction is often complicated, which makes a model-based solution elusive. We propose a learning-based approach for removing the multipath distortions for a ToF camera in a robotic arm setup. Our approach is based on deep learning. We use the robotic arm to automatically collect a large amount of ToF range images containing various multipath distortions. The training images are automatically labeled by leveraging a high precision structured light sensor available only in the training time. In the test time, we apply the learned model to remove the multipath distortions. This allows our robotic arm setup to enjoy the speed and compact form of the ToF camera without compromising with its range measurement errors. We conduct extensive experimental validations and compare the proposed method to several baseline algorithms. The experiment results show that our method achieves 55\% error reduction in range estimation and largely outperforms the baseline algorithms.
\end{abstract}

\section{Introduction}
\label{sec:intro}

Time-of-flight (ToF) cameras capture scene depth by measuring phase delay of a modulated signal emitted from an infrared LED traveling in the space. Compared to a structured light sensor or a laser scanner, ToF cameras have the advantages of low cost, high speed, and compact form. These are important system parameters for the robotic arm setup considered in the paper, which consists of a robotic arm and a range sensor. A compact sensor in the robotic arm is less likely to collide with other objects during operation. The high speed feature allows objects to remain in motion during imaging. 

\begin{figure}[t]
	\begin{center}
		 \includegraphics[trim = 0in 1in 0in 0in,width = 3.2in]{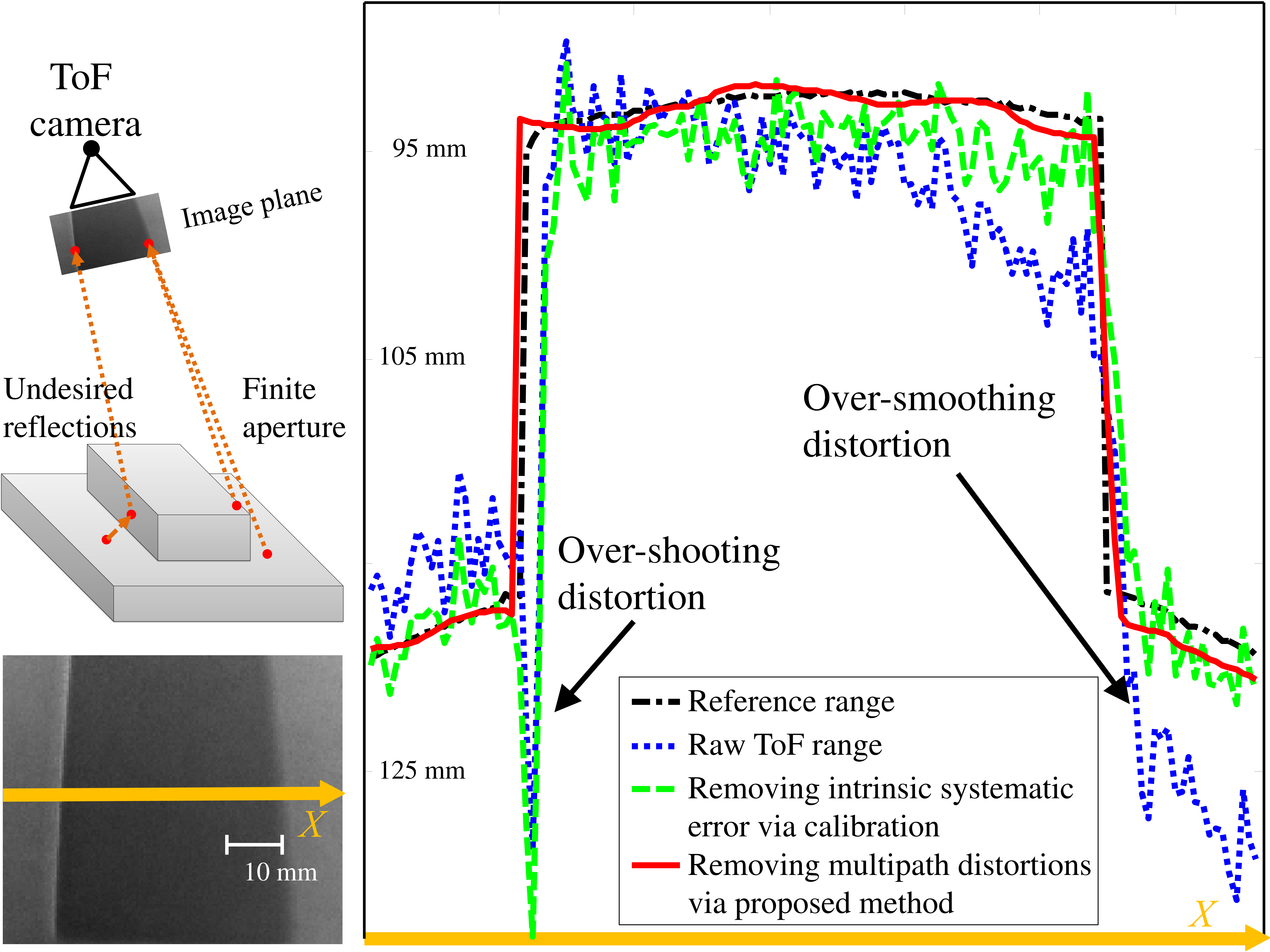}
	\end{center}
   \caption{A ToF camera captures a scene of a wood box placed on a flat platform. On the left side of the box, we observe the range over-shooting distortions (dotted blue curve)~\cite{dorrington2011separating}, which are caused by direct reflections intermingling with the indirect reflections from nearby structures. On the right side, we observe the range over-smoothing distortions, which are due to the superposition of reflection signals from the foreground and background objects. The reference range measurements (dashed black curve) are obtained from a high precision structured light sensor. We note that calibration suppresses intrinsic systematic errors but cannot correct the multipath distortions (dashed green curve). Our method learns the distortion pattern and can largely correct the multipath distortion (solid red curve).}
	\label{fig:problems}
\end{figure}

Despite the advantages, ToF cameras are not yet widely deployed in real world applications. This is mainly because range measurements from ToF cameras are error-prone. In addition to random noise, ToF range measurements also suffer from intrinsic systematic errors and multipath distortions. The intrinsic systematic errors are due to the limitations in manufacturing the ToF camera hardware, including the infrared emitter (non-ideal sinusoidal modulation), the sensor (non-identical CMOS gates), and the optics (radial distortion and vignetting effects)~\cite{SJ2008_general,BOOK2013_general}. The systematic errors result in measurement bias but can be reduced by calibration~\cite{ fuchs2008extrinsic, kim2008design, lindner2006lateral, lindner2010time}. On the other hand, the multipath distortions are the result of interaction between light signals and scenes; they include 1) range over-shooting distortions due to the superposition of the reflection light signals from nearby structures, and 2) range over-smoothing distortions due to the superposition of the reflection light signals from the foreground and background objects as illustrated in Figure~\ref{fig:problems}.

We propose a learning-based approach for removing the multipath distortions in range images captured by a ToF camera in a robotic arm setup. During training, we learn the multipath distortion patterns. The training data are automatically collected and labeled using the robotic arm and a high precision structured light sensor available only during training. After training, we apply the learned model to remove the multipath distortions in the ToF range images. This allows our robotic arm setup to enjoy the speed and compact form of the ToF camera without compromising with the range measurement errors. Our approach is based on two neural networks (NNs) where one NN learns a mapping from the ToF measurements to the true range values, while the other learns to detect object boundaries for guiding the geometry content propagation in a geodesic filtering framework~\cite{CVPR2013_enhance}. We conduct extensive experimental validations and compare our method with several baseline approaches. The results show that our method achieves more than 55\% noise reduction and largely outperforms the baseline methods.


\subsection{Related Work} 

Several methods based on ray tracing are proposed for removing multipath distortions in ToF range images~\cite{CVPR2012_multi,SPIE2012_multi,fuchs2010multipath,ICVS2013_multi}. The distortion reduction process starts with an initial estimate of the true range image, which is then gradually refined using physics-based simulation. Preliminary results along this research line are shown on simple scenes consisting of several planar objects. Extending them to complex scenes is a non-trivial task. In addition, ray tracing are computationally expensive.

Setups utilizing multiple modulation frequencies are proposed for removing the multipath distortions in~\cite{ICME2013_multi,OL2014_multi,ECCV2014_multi,GuptaTOG2015,dorrington2011separating}. These methods require special hardware and, hence, are inapplicable to off-the-shelf ToF cameras where only one modulation frequency is available. Stereo ToF camera~\cite{ICCV2011_stereo} is proposed to improve the range measurements, but it requires two ToF cameras and a baseline, which makes the setup non-compact. Our method does not require special hardware and is directly applicable to off-the-shelf ToF cameras.

Along with the range image, ToF cameras also capture an amplitude image. The amplitude image records the strength of the returning light per pixel, which quantifies the signal-to-noise ratio for each pixel. It can be used as a confidence measure. Frank~\etal~\cite{opticalEngin09} propose a filtering method where the filter shape is determined by the amplitude-based confidence measure. However, the amplitude-based confidence measure itself can be corrupted with the multipath distortions. Reynolds~\etal~\cite{CVPR2011_confidence} propose learning a confidence measure using a random forest regressor based on hand-crafted features.


Lenzen~\etal~\cite{TV2013} propose a Total Variation (TV) method for ToF range image denoising. It estimates the true range image by minimizing a predefined objective function that consists of data-fidelity terms and regularization terms. Although the method is effective in reducing random noise, it does not model multipath distortions and tends to smooth out fine structures in the range images.

Recently, deep neural networks are applied for image processing tasks including image denoising~\cite{burger2012image} and superresolution~\cite{dong2014learning}. However, since ToF range images have very different characteristics to conventional intensity images, these approaches are not directly applicable to removing the multipath distortions in ToF range images.

\subsection{Contributions} 


The contributions of the paper are summarized below.
\begin{itemize}
\item We propose a learning-based approach for removing the multipath distortions in range images captured by Time-of-Flight cameras where the training data are automatically collected and labeled using a robotic arm setup.
\item Our approach is based on geodesic filtering and two NNs where one NN learns a mapping from the ToF measurements to the true range values, while the other learns to detect object boundaries for guiding the geometry content propagation in the geodesic filtering.
\item We conduct extensive experimental validations and compare the proposed method to several baseline algorithms. The results show that the proposed method achieves 55\% noise reduction and significantly outperforms the baseline methods.
\end{itemize}

\section{Overview}
\label{sec::overview}

\begin{figure}[t]
	\begin{center}
     \includegraphics[trim =0in 2.9in 0.5in 0in , width=0.95\linewidth]{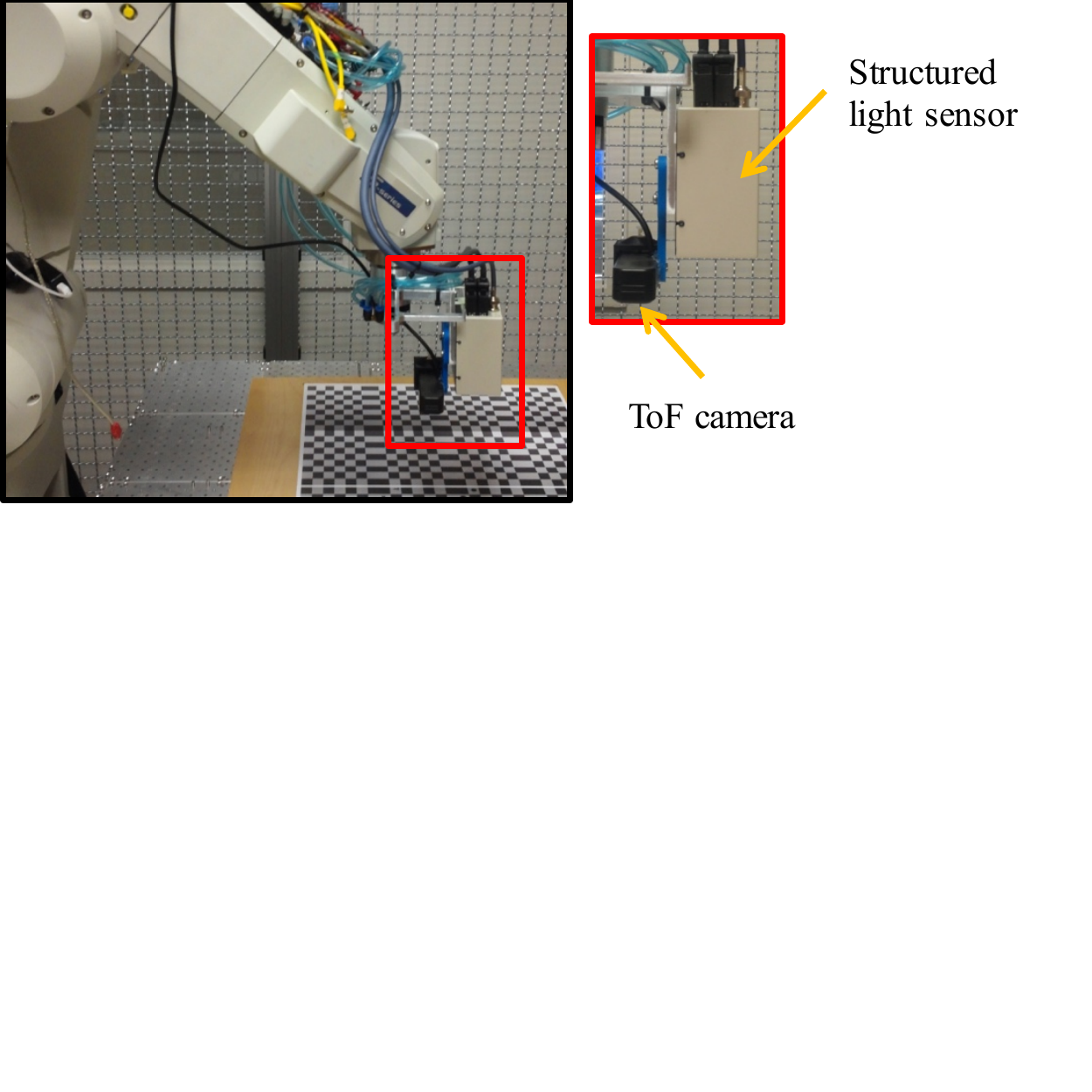}
	\end{center}
   \caption{During training, our setup consists of a ToF camera and a high precision structured light sensor mounted on a robotic arm. We perform hand-eye calibration to estimate the transformation between the two sensors, which is used for computing reference (ground truth) range values for learning. We move the robotic arm to different viewpoints to automatically collect large amount of labeled training data with different vairations. After training, we apply the learned model to remove the multipath distortions in the ToF range image.}
	\label{fig:calib_setup}
\end{figure}

We consider a setup where a ToF camera is mounted on a robotic arm as illustrated in Figure~\ref{fig:calib_setup}. In the training phase, we use a high precision structured light sensor to compute a reference (ground truth) range image for each ToF range image. We use the robotic setup to efficiently capture a large labeled training dataset for learning the model. After training, the structured light sensor is dismounted and we apply the learned model to remove the multipath distortions in the ToF range image. The setup has several benefits including:
\begin{enumerate}
\item It leverages the range measurement\footnote{In practice, we use a fusion approach that combines several range images captured by the structure light sensor for obtaining more accurate range measurments.} from the structured light sensor to compute the reference range values, avoiding expensive human annotation.
\item It can effortlessly capture a large amount of labeled data by automatically positioning the robotic arm at different viewpoints.
\item In the test phase, we directly apply the learned model for removing the multipath distortions in the ToF range image. It enjoys the compact form and high speed of the ToF camera without suffering from its error-prone range measurements.
\end{enumerate}

\section{Removing Multipath Distortions}\label{sec:scene_dependent}

\begin{figure*}[t]
	\begin{center}
     \includegraphics[trim =0in 1.5in 0in 0in , width=0.95\linewidth]{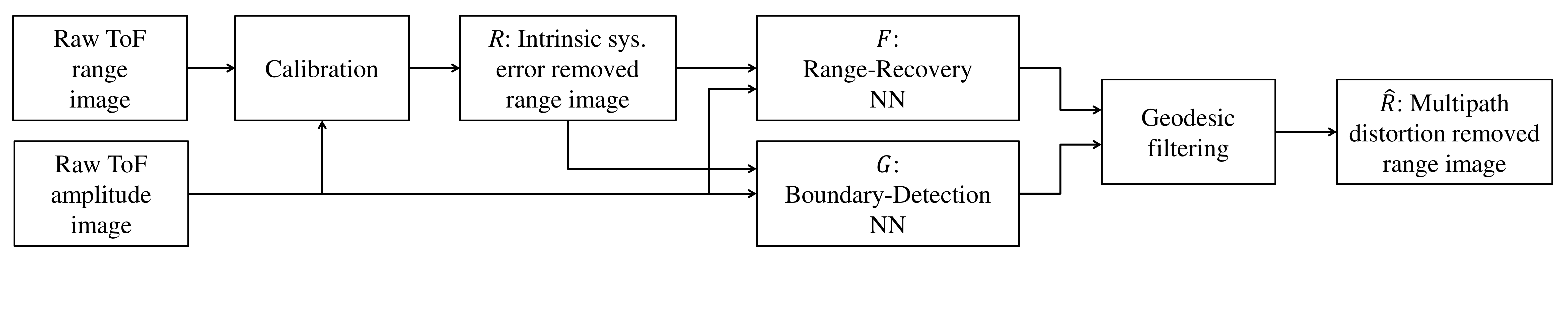}
	\end{center}
   \caption{We first apply a simple calibration model to remove intrinsic systematic errors in the ToF range image, where the result is denoted as $R$. We then use the range-recovery NN $F$ to estimate the true range values and the boundary-detection NN $G$ to detect the object boundaries. A geodesic filtering algorithm is used to compute the final range estimate $\hat{R}$.}
	\label{fig:overview}
\end{figure*}

Our method is based on feed-forward NNs. It consists of two NNs, referred to as $F$ and $G$, where $F$ defines a mapping from the ToF range measurements to the true range values, while $G$ learns to detect the true object boundary locations for guiding the geometry content propagation. Ideally, $F$ alone should be sufficient to recover the true range measurements. However, we found that although $F$ has good performance in the smooth region, it has difficulty in the object boundary region. Hence, we use $G$ in conjunction with a geodesic filtering algorithm to propagate the recovered range measurements from the smooth region to the boundary region, which improves the range estimation accuracy in the boundary region. The proposed method is summarized in Figure~\ref{fig:overview}.

The proposed method computes the final range value $\hat{R}(\mathbf{p})$ at pixel $\mathbf{p}$ as
\begin{align}
	\hat{R}(\mathbf{p}) = \dfrac{\sum_{\mathbf{q}\in \mathcal{N}_{G}(\mathbf{p})}{w_{G}(\mathbf{p},\mathbf{q})R_F(\mathbf{q})}}{\sum_{\mathbf{q}\in \mathcal{N}_{G}(\mathbf{p})}{w_{G}(\mathbf{p},\mathbf{q})}},
\label{eqn::denoise}
\end{align}
where $R_F(\mathbf{q})$ is the estimated range value at pixel $\mathbf{q}$ from applying $F$. The set $\mathcal{N}_{G}(\mathbf{p})$ denotes the neighbors of $\mathbf{p}$, and $w_{G}(\mathbf{p},\mathbf{q})$ is a weighting function measuring the similarity between pixels $\mathbf{p}$ and $\mathbf{q}$. The weighting function is given by
\begin{align}
w_{G}(\mathbf{p},\mathbf{q}) = \exp(\dfrac{-D_{G}^{2}(\mathbf{p},\mathbf{q})}{2\sigma^2}) ,
\label{eqn::weighting_func}
\end{align}
where $D_{G}(\mathbf{p},\mathbf{q})$ is the geodesic distance between the two pixels. The kernel bandwidth $\sigma$ is set to 2 in our experiments. We note that both the neighbor set $\mathcal{N}_{G}$ and the weighting function $w_{G}$ depend on $G$.

We use an edge map to compute the geodesic distances between pixels on the image grid. On the image grid, the distance between two neighboring pixels is set to a small constant if they are not in two sides of an edge; otherwise, it is set to $\infty$. The geodesic distance $D_{G}(\mathbf{p},\mathbf{q})$ is given by the length of the shortest path between the two pixels. In this way, we ensure that the pixels used to smooth a target pixel carry more weights if they are from the same surface as the target pixel. Computing pairwise geodesic distance is computationally expensive. We use the fast approximation algorithm introduced in~\cite{CVPR2013_enhance} to compute the geodesic distances from a pixel to its $K$ nearest neighbors simultaneously. The constant $K$ is set to 81 in our experiments.

We assume that the multipath distortions in the ToF range images are mostly local---the contribution to the multipath distortions from remote objects is much less than that from nearby structures. For example, the distortion in the left side of the wood box in Figure~\ref{fig:problems} is mainly from the signal around the left side corner and is non-related to the signal around the right side corner of the wood box. Such an assumption can be easily satisfied for the assembly line configuration, which is a major application target of the robotic arm setup.

Based on the assumption, we adopt a patch-based representation. We use the measurement data in the image patch centered at a pixel to construct the input data for inferring the target quantity at the pixel for both $F$ and $G$. Most of off-the-shelf ToF cameras capture an amplitude image in addition to the range image. The amplitude image represents the strength of the returning signals at each pixel. Both raw range and amplitude images are inputs to the NNs.

\subsection{Range-Recovery NN} 

We learn a regression function that maps the ToF measurement data in the patch centered at the target pixel location to its range value. Let $\mathbf{p}$ be the target pixel. The input data for each pixel is given by
\begin{align}
[\mathbf{x}_{R}(\mathbf{p})^T \medspace \mathbf{x}_{A}({\mathbf{p}})^T\medspace \mathbf{b}_{R}(\mathbf{p})^T \medspace \mathbf{b}_{A}(\mathbf{p})^T].
\label{eqn::input_encoding}
\end{align}
The term $\mathbf{x}_{R}(\mathbf{p})$ is a vector encoding the range values of the pixels in the patch centered at $\mathbf{p}$. It is obtained by subtracting the range values by the range value of $\mathbf{p}$, followed by a normalization step, mapping the values to $[-0.5 \medspace 0.5]$. The second term, $\mathbf{x}_{A}({\mathbf{p}})$, is obtained via applying a similar operation to the amplitude values in the patch. The patch size used in our experiments is $11\times 11$. Hence, the dimensions of $\mathbf{x}_{R}(\mathbf{p})$ and $\mathbf{x}_{A}({\mathbf{p}})$ are both 120.

The last two terms in~(\ref{eqn::input_encoding}), $\mathbf{b}_{R}(\mathbf{p})$ and $\mathbf{b}_{A}(\mathbf{p}$), are binary vectors encoding range and amplitude values of $\mathbf{p}$, which are subtracted to compute $\mathbf{x}_{R}(\mathbf{p})$ and $\medspace \mathbf{x}_{A}({\mathbf{p}})$. The dimensions of the binary vectors are set to 20 in our experiments. They are computed by uniformly quantizing the range and amplitude values into $20$ different intervals, respectively. We set the corresponding element of the binary vector to $1$ if the range and amplitude values fall in the interval; otherwise 0.

The target value $t$ for the range-recovery NN is given by the difference between the ground truth range value and the ToF range value; that is
\begin{align}
t({\mathbf{p}}) = R^{*}({\mathbf{p}})-R({\mathbf{p}}),
\end{align}
where $R^{*}({\mathbf{p}})$ and $R({\mathbf{p}})$ are the ground truth and the ToF range value at $\mathbf{p}$, respectively. We truncate the target value $t$ using a threshold, which is 15 mm in our experiments. The pair $(\mathbf{x}({\mathbf{p}}), t(\mathbf{p}))$ serves as the training data for learning $F$.

The range-recovery NN is defined as a three-layer feed-forward network where the first layer contains $40$ neurons, and the second and third layer each contains $10$ neurons. The neurons are fully connected. We apply rectified linear units (ReLU) after the neurons for modeling the non-linear relationship between the input data and target value. We do not use the pooling layer popular in modern convolutional neural networks since pooling reduces the output resolution. The neurons in the third layer are fed into the network output neuron. We train $F$ through minimizing the Euclidean loss. We use the learned $F$ to recover the true range value. Note that the recovered range is given by 
\begin{equation}
R_F(\mathbf{p}) = F(\mathbf{p})+R({\mathbf{p}}).
\end{equation}

\subsection{Boundary-Detection NN} 

We compute the ground truth boundaries and their directions by applying the Canny edge detector to the reference range images obtained with the structured light sensor. We divide the detected edges into $4$ groups based on a uniform quantization of the edge orientation. For each group, the edges from the other groups as well as non-edge pixels are used as the negative training samples. They are used to learn $4$ edge detectors, each for one direction.

We use $4$ separate feed-forward NNs for the $4$ edge detectors. Similar to the case of the range-recovery NN, we extract the measurements surrounding a pixel $\mathbf{p}$ within a patch to form an input vector 
\begin{align}
[\mathbf{x}_{R}(\mathbf{p})^T \medspace \mathbf{x}_{A}({\mathbf{p}})^T].
\end{align}
The structures for the $4$ boundary-detection NNs are identical. Each consists of two layers where the first layer has 40 neurons and the second layer has 20 neurons. The neurons are all fully connected. Each layer is followed by the ReLU nonlinearity. The output layers have two units, which represent the edge and non-edge likelihood scores after applying a softmax operation. The networks are trained via minimizing the cross entropy loss.

For each pixel, we find a maximum response from the $4$ NNs for the boundary likelihood score and direction, which is similar to the procedure described in~\cite{amfm_pami2011}. We apply the non-maximum suppression and hysteresis thresholding to compute the binary edge map for geodesic filtering as used in the Canny edge detector.

\section{Datasets}
\label{sec:dataset}

\begin{figure}[t]
	\begin{center}
		 \includegraphics[trim = 0in 0.5in 0.7in 0in,width=0.99\linewidth]{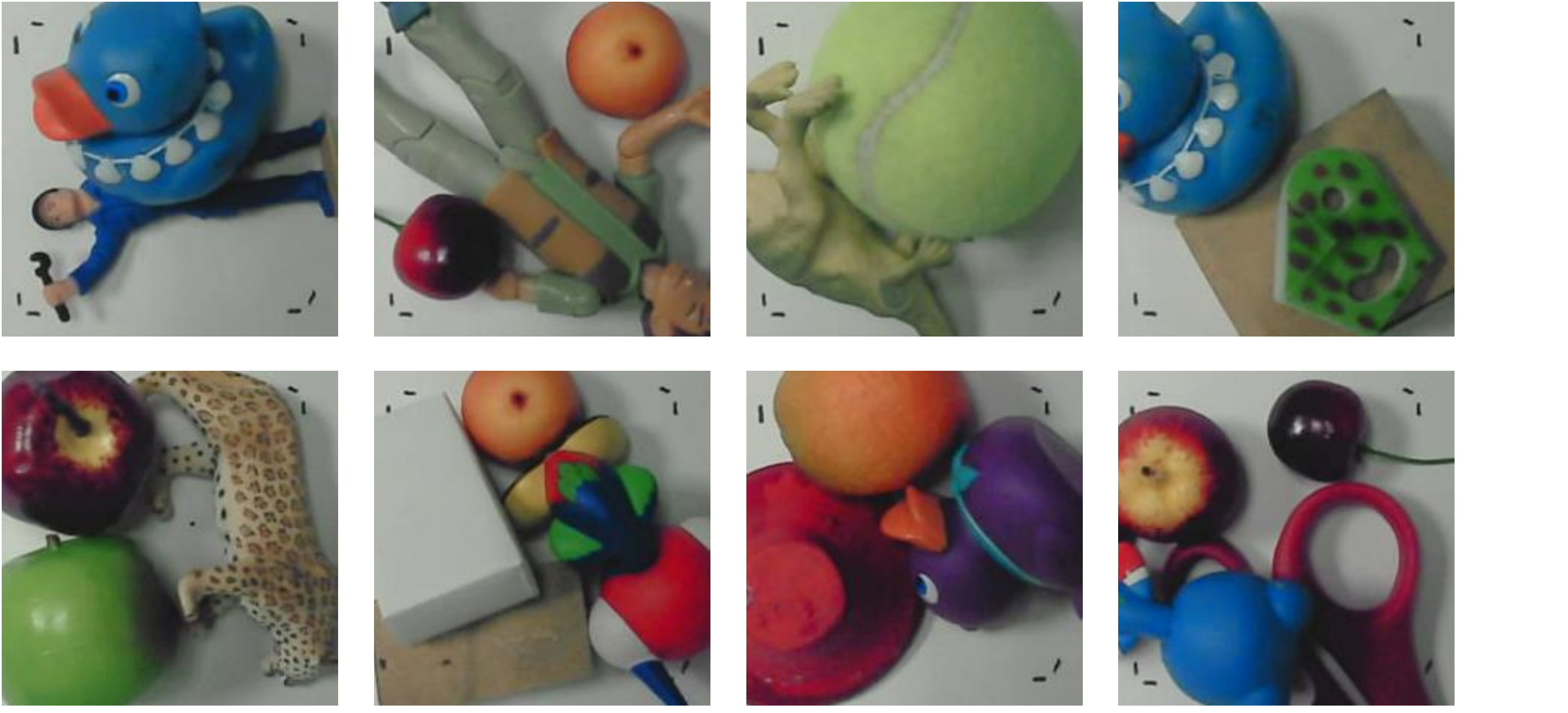}
	\end{center}
   \caption{The images illustrate the scenes used for training our model for removing multipath distortions in our experiments.}
	\label{fig::training_scene}
\end{figure}

\begin{figure*}[tb!]
	\begin{center}
		 \includegraphics[trim =1.2in 3.2in 1.0in 3in,width=0.39\linewidth]{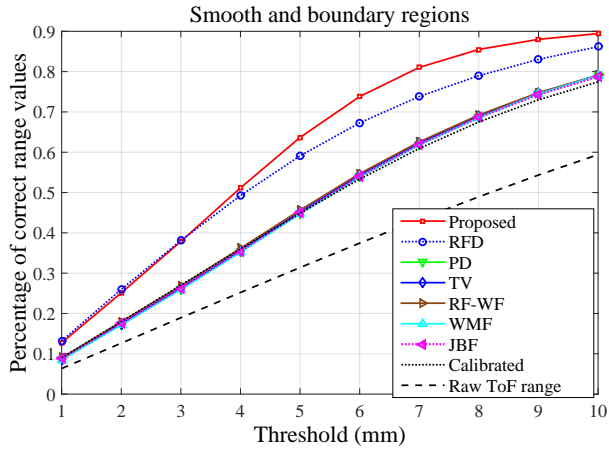} \hspace{0.7in}
		 \includegraphics[trim =1.2in 3.2in 1.0in 3in, width=0.39\linewidth]{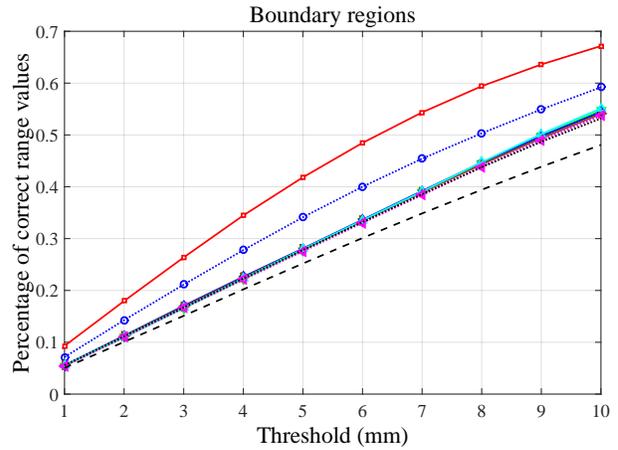}
	\end{center}
   \caption{{Quantitative comparison.} The left plot shows the distortion removal performance in both the smooth and boundary regions, while the right plot shows the distortion removal performance in the boundary regions.}
	\label{fig::denoising_performance_comparison}
\end{figure*}

We used the robotic arm setup discussed in Section~\ref{sec::overview} to obtain labeled range images. We collected two datasets, one for training and the other for performance evaluation, which are called MD-Train and MD-Test, respectively\footnote{The MD (Multipath Distortion) datasets are available in \url{https://sites.google.com/site/kilhoson/learning_tof}}. The MD-Train contained 30 scenes, which was constructed by randomly arranging a subset of 30 different objects on a table. The dimensions of the objects were between 100 to 200 mm. We visualize some of the scenes in Figure~\ref{fig::training_scene}. For each scene, we captured ToF images from 18 different viewpoints, which constituted 540 training images. During data capture, our ToF camera was about 180 to 350 mm above the objects, which means that our objects appear relative large, and the captured ToF range images contain significant amount of multipath distortions as shown in Figure~\ref{fig::denoising_performance_comparison}.

The precision of the structured light sensor was 0.5 mm. We performed the hand-eye calibration before the dataset collection. Hence, the coordinate transformation between the structured light sensor and the ToF camera was known. We used a measurement fusion approach to compute the reference range image for each ToF range image. Due to the baseline between the projector and camera, the structured light sensor failed to obtain the range measurements in the occluded area. This problem was mitigated by scanning the scene with 10 different in-plane rotation angles at 3 different distances from the objects. We transformed all the 30 scans to the ToF camera coordinate system. For each pixel in the ToF camera, we considered the 3D points projected to the pixel, found the cluster of 3D points closest to the camera center, and used the median of the range values of the 3D points in the cluster as the reference range value. If an insufficient number of 3D points were projected to a pixel, the reference range for the pixel was denoted as missing. This happened because the field of views of the sensors were different and because some regions were occluded even using the multiple viewpoints. The missing pixels were marked in the green color in the figures reported in the experiment section. The missing pixels were neither used for training nor for performance evaluation.

The MD-Test was captured in the same way. It contained 20 scenes constructed by randomly arranging a subset of 30 different objects in the platform. Note that none of the objects used in the MD-Train was used in the MD-Test. For each scene, we also captured ToF images from 18 different views, providing 360 images. We used the structured light sensor to compute the reference range image for the purpose of performance evaluation.

\section{Experiments}
\label{sec::expr}

\begin{figure*}[tb!]
	\centering
		 \includegraphics[trim =0in 0.8in 0.5in 0in, width=1\linewidth]{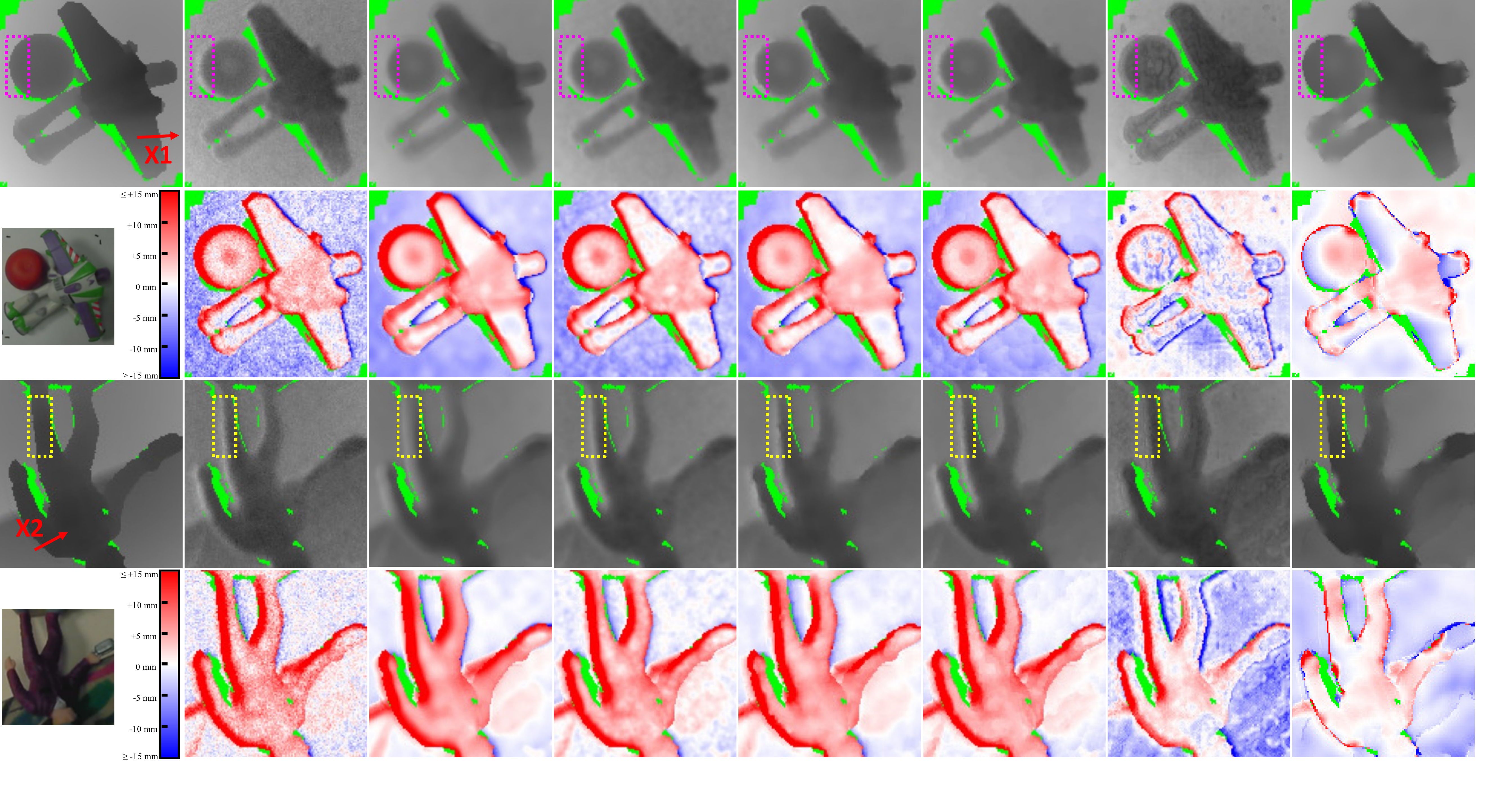}
		\begin{tabular}{cccccccc}
 (a) & \hspace{1.35cm} (b) & \hspace{1.35cm}(c) & \hspace{1.35cm}(d) & \hspace{1.35cm}(e) & \hspace{1.35cm}(f)& \hspace{1.35cm}(g) & \hspace{1.35cm}(h)\\		
 		\end{tabular}
		\caption{{Qualitative comparison.} (a)~Reference range images and scenes. The 1st and 3rd rows show results of (b)~intrinsic systematic error calibrated, (c)~JBF~\cite{jbf_CVPR2007}, (d)~RF-WF~\cite{opticalEngin09, CVPR2011_confidence}, (e)~PD~\cite{eccv2012_enhance}, (f)~TV~\cite{TV2013}, (g)~RFD~\cite{CVPR2011_confidence}, and (h)~the proposed algorithm. (Due to the space limitation, the visual result of WMF is not displayed in this Figure.) The 2nd and 4th rows visualize the difference between the reference range image and the results of the algorithms. The errors are color-coded. Pixels in the white color region have correct range values. The green region denotes the place where reference range measurements are missing.}
		\label{fig::qualitative}
\end{figure*}

Our setup can be applied to a wide variety of off-the-shelf ToF cameras, such as those from Mesa Imaging, PMD, and SoftKinetic. In this paper, we used a SoftKinetic DS325 ToF camera, having a resolution of 320$\times$240\footnote{The SoftKinetic DS325 ToF camera also has a color camera. However, it was used only for visualization purpose in Figures~\ref{fig::training_scene} and~\ref{fig::qualitative} and not for computation.}. In the experiments, we only used the central portion of the ToF image with a resolution of 180$\times$140. We placed a neutral density filter in front of the infrared LED. It attenuated the flash and allowed us to range close objects. We applied a calibration model to remove the intrinsic systematic error before any processing. The calibration method defined a linear model for each pixel location for relating the observed measurements (both range and amplitude) to the true range measurements. We captured a checkerboard pattern (shown in Figure~\ref{fig:calib_setup}) with 770 different poses to fit the linear models. The intensities of the checkerboard pattern were gradually increased from top to bottom (from 0\% to 50\% for the darker patches, from 50\% to 100\% for the brighter patches). This increased the variety of the amplitude measurements and empirically rendered better intrinsic systematic error calibration performance. The raw ToF range images were calibrated before applying the proposed algorithm as well as before applying the baseline algorithms for comparison.

The 540 training images in the MD-Train generated about 10 million image patches. Hence, our NNs learned the input-output relationship from the 10 million training samples. We used the public Caffe library~\cite{jia2014caffe} to train our NNs. Stochastic gradient descent with momentum was used for optimization. The momentum weight was set to 0.9, while the batch size was set to 500. We trained the NNs for 40 epochs.

\begin{figure*}[t]
	\centering
		 \includegraphics[trim =0in 0.4in 0in 0in, width=0.95\linewidth]{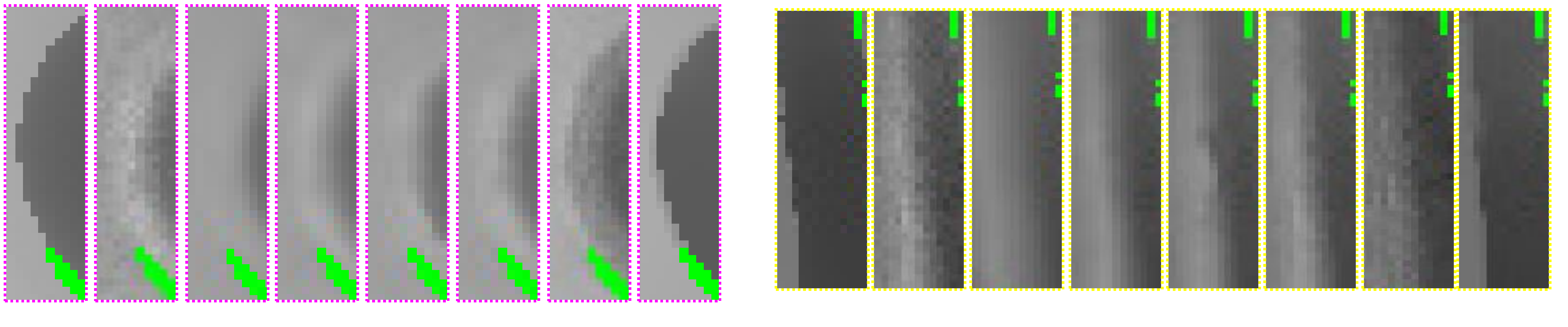}
		\caption{Close ups of the range values in the magenta and yellow rectangle regions in Figure~\ref{fig::qualitative}. The column order is the same as that in Figure~\ref{fig::qualitative}. The proposed algorithm outperforms the competing algorithms in removing distortions in the boundary regions.}
		\label{fig::detailed}
\end{figure*}
\begin{figure*}[t]
	\centering{
		 \hspace{0.1in}%
		\includegraphics[width=0.46\linewidth]{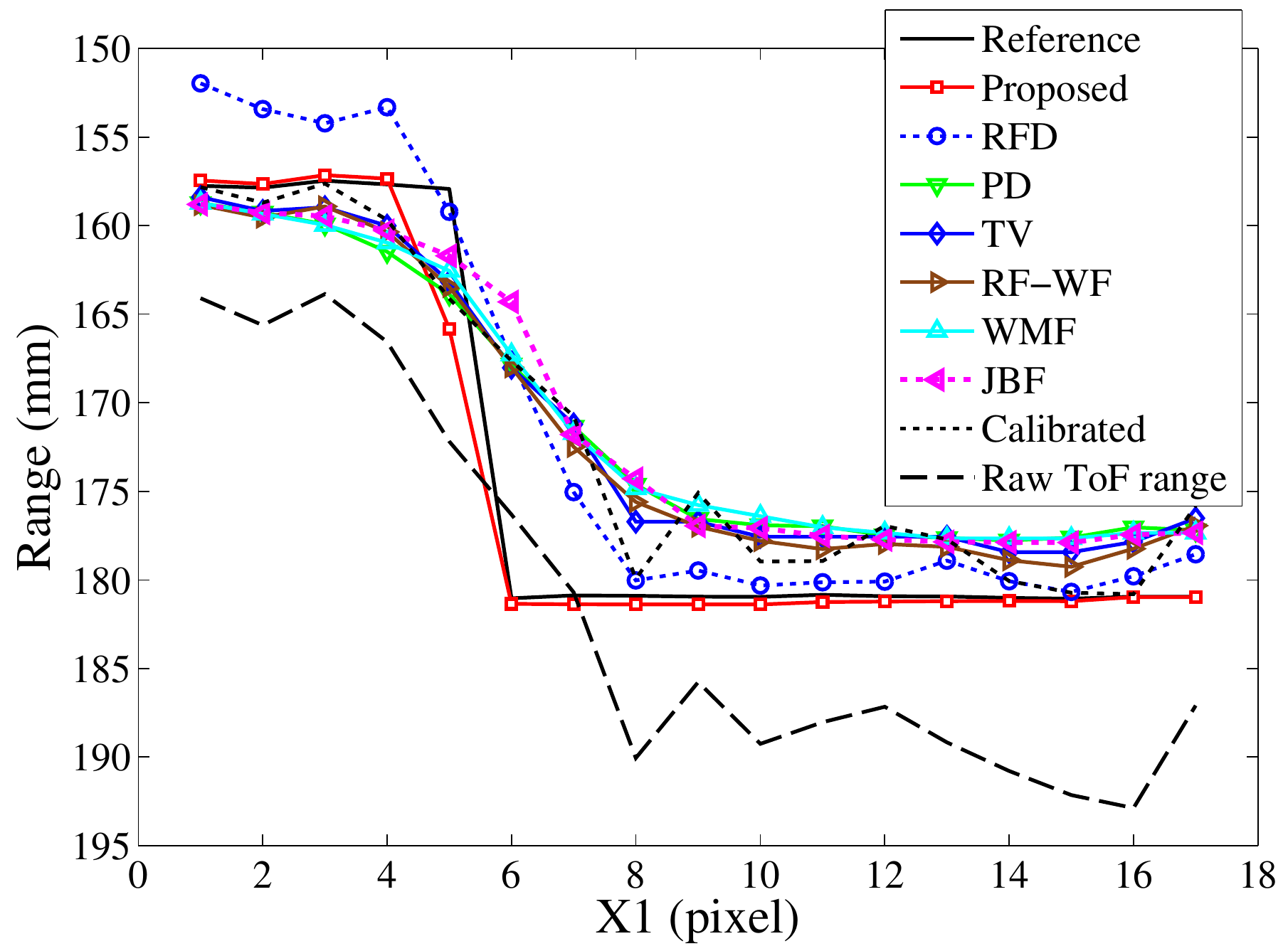}\hspace{0.4in}
		\includegraphics[width=0.46\linewidth]{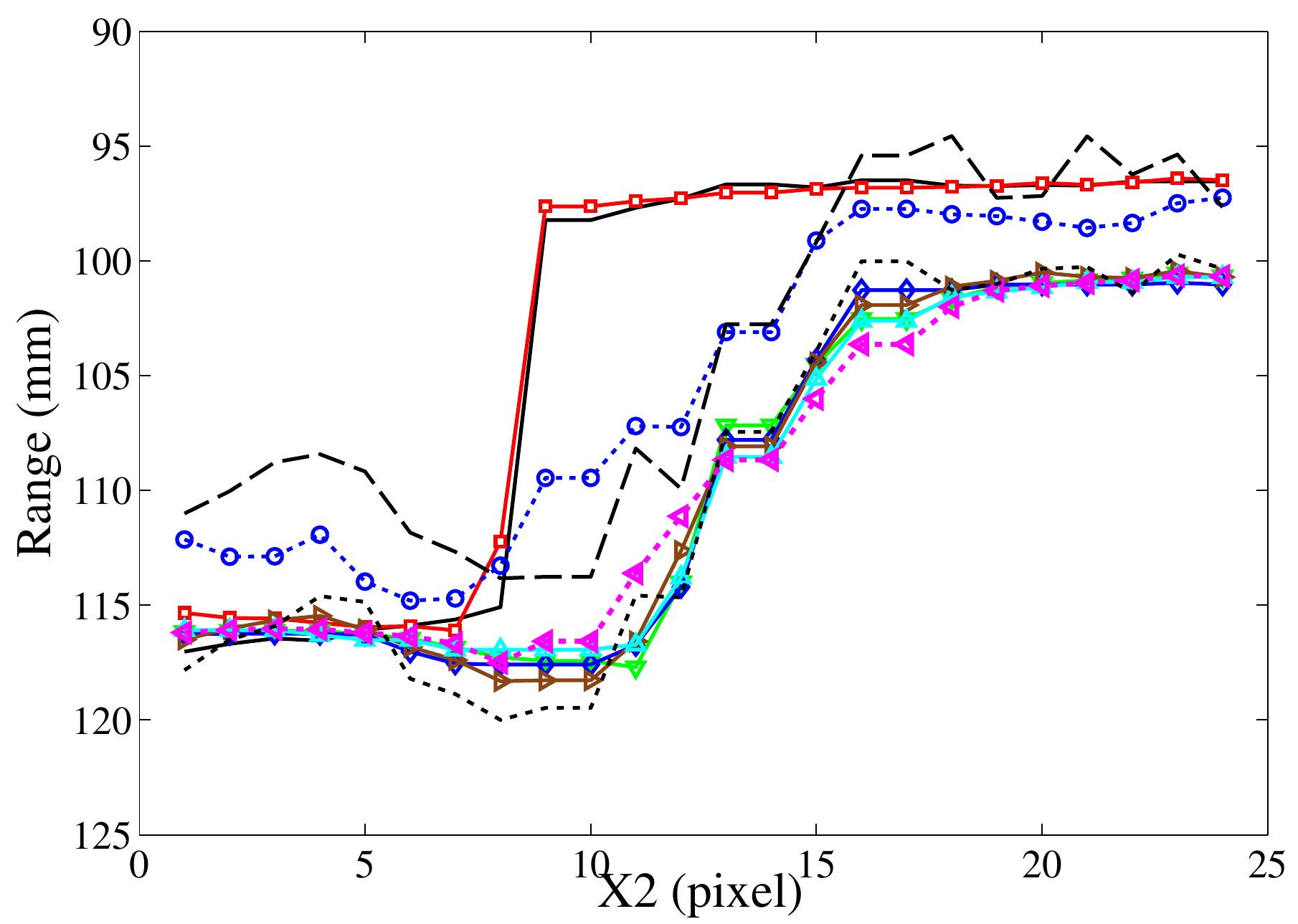}}
		\caption{Range values along the red arrows X1 and X2 in Figure~\ref{fig::qualitative}. The ToF range values suffer from multipath distortions, which are largely corrected by the proposed algorithm.}
		\label{fig::boundary_profile}
\end{figure*}

\subsection{Baseline Methods} 

We compared our algorithm with several competing algorithms for removing ToF range distortions, including 
\begin{enumerate}
\item {\bf JBF:} The ToF amplitude-guided joint bilateral filtering algorithm~\cite{jbf_SIG2004, jbf_CVPR2007, jbf_pami2013}.
\item {\bf WMF:} The ToF amplitude-based weighted median filtering method~\cite{wmf_tip2013, wmf_ICCV2013,wmf_cvpr2014}. 
\item {\bf RF-WF:} The random forest regression-based weighted filtering method~\cite{opticalEngin09, CVPR2011_confidence}.
\item {\bf RFD:} The random forest regression-based distortion removal algorithm~\cite{CVPR2011_confidence}.
\item {\bf TV:} The total variation-based algorithm~\cite{TV2013}.
\item {\bf PD:} The shape-prior based patch algorithm~\cite{eccv2012_enhance}.
\end{enumerate}

\begin{figure*}[t]
	\begin{center}
		 \includegraphics[trim =1.5in 5.5in 1.2in 1.0in, width=0.42\linewidth]{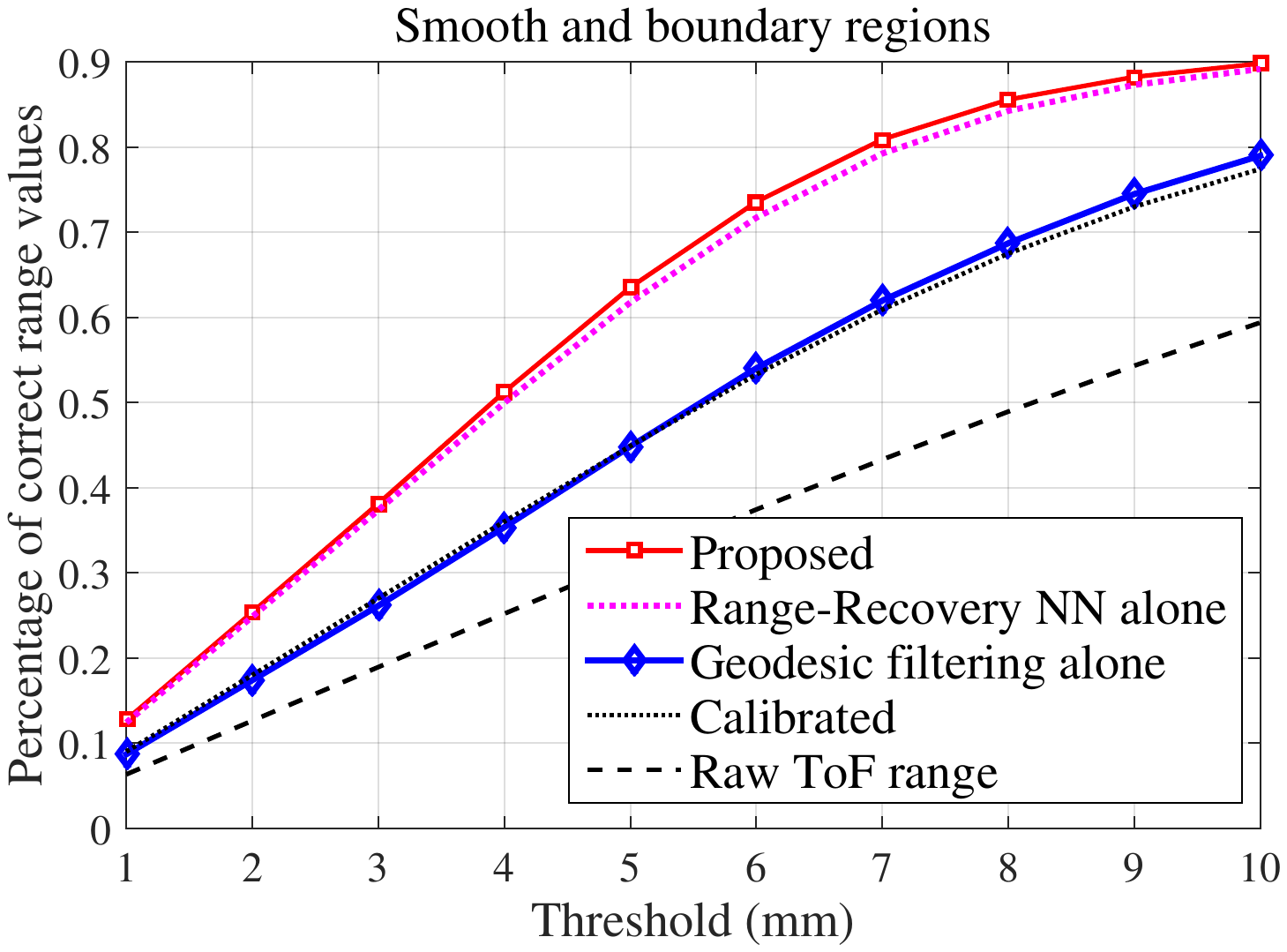}\hspace{0.5in}
		 \includegraphics[trim =1.5in 5.5in 1.2in 1.0in, width=0.42\linewidth]{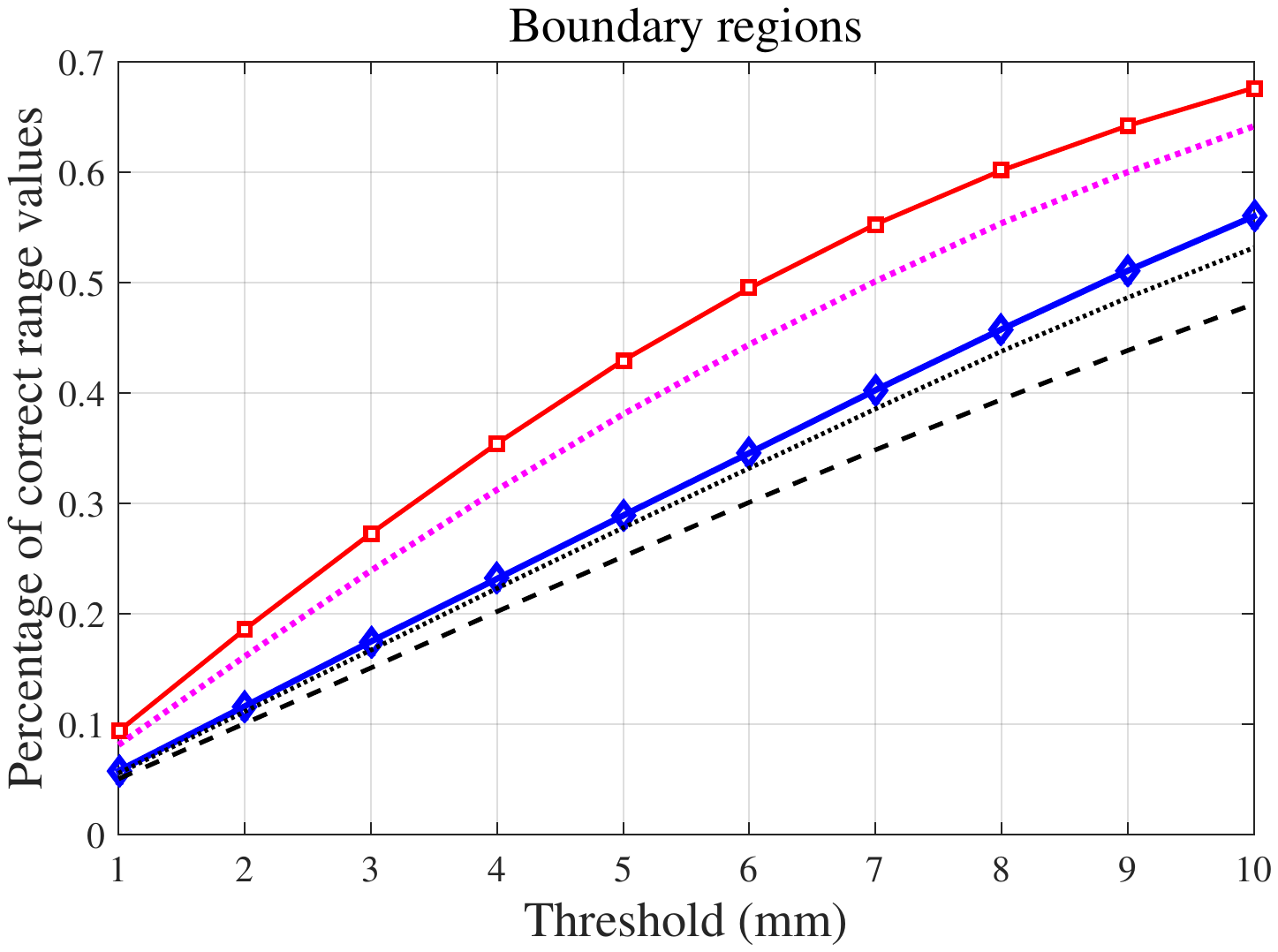}
	\end{center}
   \caption{Algorithm component analysis, demonstrating the contribution of individual components in the proposed algorithm.}
	\label{fig::component_analysis}
\end{figure*}

The amplitude image was used in the JBF and WMF as the guidance image. The RF-WF used a weighted Gaussian filter. The weights in the RF-WF were determined by a random forest regressor. Note that \cite{CVPR2011_confidence} proposed using the random forest regressor to learn a confidence measure but did not consider removing range distortions in ToF range images. We applied the learned confidence measure from the random forest regressor~\cite{CVPR2011_confidence} with the weighted filtering method proposed~\cite{opticalEngin09} to remove the multipath distortions in the ToF range image. We used the MD-Train dataset to train the random forest regressor. For the RFD, the random forest regressor~\cite{CVPR2011_confidence} was directly trained to remove the distortion directly using the same training data as the proposed method. Comparison with the RFD justified the use of NNs for removing multipath distortion. For the TV~\cite{TV2013}, we used L1 norm for the regularization term in the objective function to preserve structure of the range image. The TV objective function was optimized by a primal-dual approach~\cite{primal_dual}. The PD algorithm~\cite{eccv2012_enhance} was proposed for the ToF range image superresolution task. It formulated the superresolution as an inference problem in the Markov Random Field where the latent variables were the indices to the reference patches. The data term was the L2 distance between observations and the estimated patches, and the pairwise term was the difference between neighboring patches. We adapted the PD algorithm for the ToF range image distortion removal task. The performance of PD depended on the comprehensiveness of the training patches; however, the computational time also increased with the number of training patches. We used the pre-generated patches provided by the authors, which were designed to be generic for various scenes.

We used the implementations provided by the authors of the original papers whenever available. The competing algorithms all had free parameters. We performed grid search and used the best one to report the performance.

\subsection{Performance Metric} 

We used the MD-Test for performance evaluation. The performance was measured by the percentage of correctly recovered range values. A range value was considered correctly recovered if its deviation from the reference range value was less than a threshold. We varied the threshold and plotted the performance curve. We computed the results for the whole image regions and for the boundary regions separately (5-pixel margin from the ground truth boundary pixels). A similar performance metric was used in the Middlebury stereo benchmark~\cite{scharstein2002taxonomy}. 

\subsection{Results}

Figure~\ref{fig::denoising_performance_comparison} compares the performance of the competing algorithms. The raw ToF range images were noisy. For the threshold of 5mm, only about 30\% of range measurements were correct. After applying the calibration for removing the intrinsic systematic error, the accuracy improved to 45\%. The TV, PD, RF-WF, WMF, and JBF algorithms improved the measurements based on the results after the intrinsic calibration. However, the improvements were marginal. The RFD was ranked 2nd, which showed the benefit of a learning-based algorithm for distortion removal. The proposed algorithm  modeled the multipath distortions using the high capacity NNs and achieved the best performance. It rendered an accuracy of 63\% at 5mm threshold, corresponding to a 55\% improvement with respect to the raw measurements.

Figure~\ref{fig::qualitative} shows qualitative comparisons. We observed that the competing algorithms were able to filter out the random noise but failed to correct the multipath distortions in the boundary regions. The results still contained severe range over-shooting errors in the boundary regions as indicated by the red color regions in the figure. In contrast, the proposed algorithm removed a large portion of the multi-path distortions. Detailed comparisons in the boundary regions are provided in Figures~\ref{fig::detailed} and~\ref{fig::boundary_profile}. We found that our algorithm largely removed the error in the boundary regions, while the competing algorithms tended to blur the boundaries.

\subsection{Algorithm Component Analysis}

Our method has two components: 1) the range-recovery NN and 2) the geodesic filtering using the boundary map computed by the boundary-detection NN. We analyzed the contributions from the two components individually using the MD-Test. Figure~\ref{fig::component_analysis} shows the performance curves of the results in the whole image and boundary regions. We found that a large portion of performance improvements were due to the range-recovery NN. Using the geodesic filtering alone led to slight improvement. However, the improvement were mostly in the boundary regions, which complimented the results from the range-recovery NN. Combining the two components, our method achieved large performance boost in both smooth and boundary regions.


\subsection{Performance of Boundary-Detection NN}

\begin{figure}[t]
	\begin{center}
    \includegraphics[trim=0in 0.5in 0.3in 0in,width=0.99\linewidth]{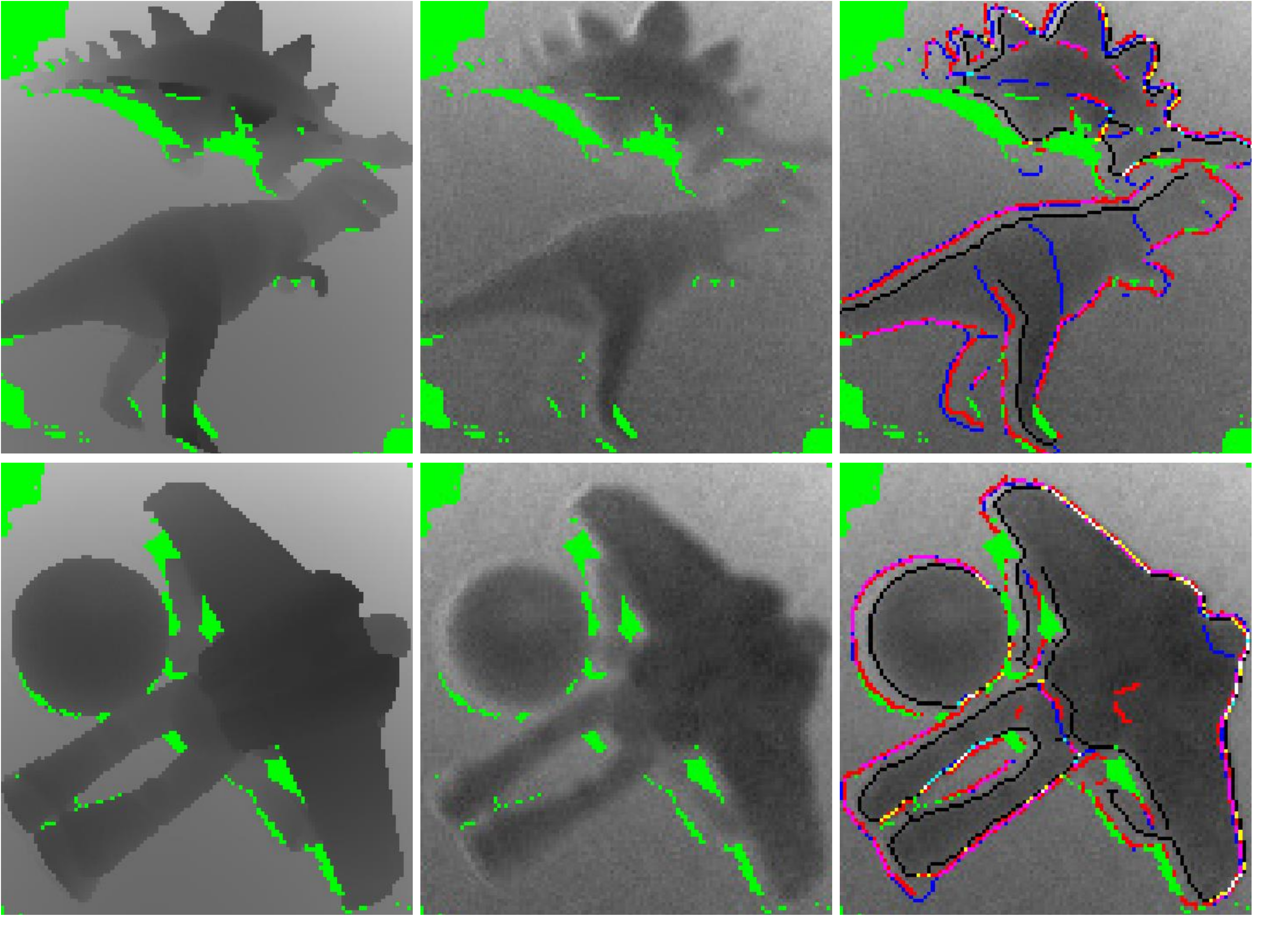}
		\begin{tabular}{c c c}
		(a) & \hspace{2cm}(b) & \hspace{2cm}(c)
		\end{tabular}
	\end{center}
   \caption{Qualitative comparison of edge detectors. (a)~Reference range images. (b)~ToF range images. (c)~Edge detection results (blue: reference edges; black: Canny edges; red: our edges; magenta: overlaps of reference edges and our edges; yellow: overlaps of our edges and Canny edges; cyan: overlaps of reference edges and Canny edges; white: overlaps of all the edges). The Canny edges have considerable drifts (5 pixels) due to the severe multipath distortions around boundary regions. In contrast, our edges are close to the ground truth edges.}
	\label{fig:edge_results}
\end{figure}

\begin{figure}[t]
	\begin{center}
     \includegraphics[trim=1.5in 6.0in 1.0in 2.5in,width=.99\linewidth]{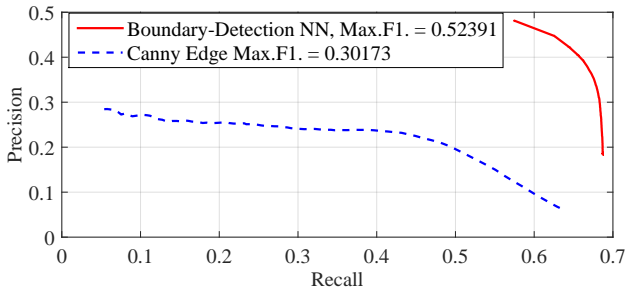}
	\end{center}
   \caption{Quantitative comparison of edge detectors, showing the precision and recall curves with the maximum F1-score.}
	\label{fig:edge_graph}
\end{figure}

Figure~\ref{fig:edge_results} compares boundary detection results by our method (red) and by the Canny detector (black). The edges detected by the Canny detector were off (5 pixels in average error) from the reference edge locations. Edges obtained with our method were much closer to the reference (blue) edges. They often exactly overlapped with each other (magenta), whereas the Canny edges were usually far away from the ground truth. Figure~\ref{fig:edge_graph} shows precision and recall curves of the edge detectors evaluated with the MD-Test, which demonstrates that our edges were significantly more accurate than the Canny edges. The accurate edge localization can be useful for edge-based object pose estimation algorithms such as~\cite{tuzel2014learning,choi2012voting}.

\subsection{Time}

The time required for training each NN was about 20 minutes. The computation of the range-recovery NN was about 35 ms using a GPU card, while that of the boundary-detection NN took about 30 ms. The geodesic filtering algorithm was performed on CPU, which took about 100 ms. The overall computation time was 165 ms. The experiments were conducted on a 4-core Intel i7 processor with an Nvidia Titan graphic card.

\section{Conclusion}

We introduced a learning-based approach for removing the multipath distortions in ToF range images, which is based on deep learning and a geodesic filtering algorithm. Experiment results showed that our approach effectively reduced the range over-shooting and range over-smoothing distortions. ToF range measurements are also known to correlate with object material property. In future, we plan to study the feasibility of using the learning-based approach for addressing the material property dependency.

\section*{Acknowledgments}

The authors would like to thank Jay Thornton, John Barnwell, and Yukiyasu Domae for their support and feedback.

\bibliographystyle{ieee}
\bibliography{tof}

\end{document}